\documentclass[runningheads]{llncs}

\usepackage[sort,square,numbers]{natbib}

\usepackage{times}
\usepackage{soul}
\usepackage{url}
\usepackage[hidelinks]{hyperref}
\usepackage{graphicx}
\usepackage{amsmath}
\usepackage{booktabs}
\usepackage[ruled,vlined,linesnumbered]{algorithm2e}
\usepackage{multirow}
\usepackage[inline]{enumitem}

\usepackage{xspace}
\newcommand{\framework}{\fsl{Noe}\xspace}

\usepackage{mas-soc}

\usepackage{amssymb}
\newlist{todolist}{itemize}{2}
\setlist[todolist]{label=$\square$}
\usepackage{pifont}

\usepackage{pgfplots}
\pgfplotsset{compat=newest}
\usepgfplotslibrary{groupplots}
\usepgfplotslibrary{statistics}

\usepackage{tikz}

\pgfplotsset{
    discard if not/.style 2 args={
        /pgfplots/boxplot/data filter/.code={
            \edef\tempa{\thisrow{#1}}
            \edef\tempb{#2}
            \ifx\tempa\tempb
            \else
                
            \fi
        }
    }
}

\SetKwInput{KwInput}{Input}                
\SetKwInput{KwOutput}{Output}              
\SetKwFunction{FMain}{Action Selection}

\definecolor{revision}{rgb}{0.29, 0.0, 0.51}
\newcommand{\revised}[1]{#1}
\newcommand{\revisedst}[1]{}

\begin{document}
\sloppy
\title{\framework: Norm Emergence and Robustness Based on Emotions in Multiagent Systems}

\author{Sz-Ting Tzeng \inst{1} \and 
Nirav Ajmeri\inst{2} \and 
Munindar P. Singh\inst{1}}
\authorrunning{ST Tzeng et al.}
%
\authorrunning{ST Tzeng et al.}
%
\institute{
North Carolina State University, Raleigh NC 27695, USA\\
\email{\{stzeng,mpsingh\}@ncsu.edu}\\
\and
University of Bristol, Bristol, BS8 1UB, UK\\
\email{nirav.ajmeri@bristol.ac.uk}\\
}



\maketitle

\begin{abstract}
Social norms characterize collective and acceptable group conducts in human society. 
Furthermore, some social norms emerge from interactions of agents or humans. 
To achieve agent autonomy and make norm satisfaction explainable, we include emotions into the normative reasoning process, which evaluates whether to comply or violate a norm. 
Specifically, before selecting an action to execute, an agent observes the environment and infers the state and consequences with its internal states after norm satisfaction or violation of a social norm. 
Both norm satisfaction and violation provoke further emotions, and the subsequent emotions affect norm enforcement. 
This paper investigates how modeling emotions affect the emergence and robustness of social norms via social simulation experiments. 
We find that an ability in agents to consider emotional responses to the outcomes of norm satisfaction and violation (1) promotes norm compliance; and (2) improves societal welfare.

\end{abstract}


\section{Introduction}
Humans, in daily life, face many choices at many moments, and each selection brings positive and negative payoffs. 
In psychology, decision-making \cite{Simon1960new} is a cognitive process that selects a belief or a series of actions based on values, preferences, and beliefs to achieve specific goals. 
Emotions, the responses to internal or external events or objects, can involve the decision-making process and provide extra information in communication \cite{Keltner1999social,Schwarz2000emotion}. 
Social norms describe societal principles between agents in a multiagent system. 
While social norms regulate behaviors in society \cite{TIST-13-Governance,Savarimuthu+Cranefield-11,TOSEM-20:Desen}, 
humans and agents have the capacity to deviate from norms in certain contexts. 
For instance, people shake hands normally but deviate from this social norm during a pandemic.
Chopra and Singh \cite{WWW-16:IOSE} describe how social protocols rely on a foundation of norms though they do not discuss how the appropriate norms emerge.

An agent that models the emotions of its users and other humans can potentially behave in a more realistic and trustworthy manner. 
The decision-making process for humans or agents involves evaluating possible consequences of available actions and choosing the action that maximizes the expected utility \cite{Edwards1954theory}.
Herbert Simon, one of the founders of AI, emphasized that general thinking and problem-solving must incorporate the influence of emotions \cite{Simon1967motivational}.
Without considering emotions or other affective characteristics, such as personality or mood, some compliance seems irrational \cite{Argente2020normative}.
Humans' compliance shows hints on rational planning over their objectives \cite{Keltner1999social}.
Including emotion or personality in normative reasoning makes these compliance behaviors explainable. 
Norms either are defined in a top-down manner or emerge in a bottom-up manner \cite{Savarimuthu+Cranefield-11,Morris2019norm}. 
Works on norms include norm emergence based on the prior outcome of norms, automated run-time revision of sanctions \cite{DellAnna+20:sanctions}, or considering various aspects during reasoning \cite{AAMAS-20:Elessar,IJCAI-18:Poros}. 
However, sanctions in the real world are often subtle instead of harsh punishments. 
For instance, sanctions could be trust updates or emotional expression and might change one's behavior \cite{Nardin+16:Sanctioning,Bourgais2019ben}. 
Kalia {\etal} \cite{IJCAI-19:Emotions} considered norm outcome with respect to emotions and trust and goals. 
Modeling and reasoning about emotions and other affective characteristics in an agent then become important in decision making and would help the agent enforce and internalize norms.

%
Accordingly, we propose \framework, an agent architecture that integrates
decision-making with normative reasoning and emotions.
We investigate the following research question. 
\begin{description}

\item[RQ\fsub{emotion}.] How does modeling the emotional responses of agents to the outcomes of interactions affect norm emergence and social welfare in an agent society?

\end{description}

To address RQ\fsub{emotion}, we refine the abstract normative emotional agent architecture \cite{Argente2020normative} and investigate the interplay of norms and emotions. 
We propose a framework \framework based on BDI architecture \cite{Rao1991modeling}, norm life-cycle \cite{Savarimuthu+Cranefield-11,Frantz2018modeling,Argente2020normative}, and emotion life-cycle \cite[pp.~62--64]{Alfonso2017agents} \cite{Marsella2009ema}.
To evaluate \framework, we design a simulation experiment with various agent societies.
We investigate how norms emerge and how emotions in normative agents influence social welfare. 

To make the problem tractable, we apply one social norm in our simulation and simplify the emotional expression to reduce the complexity. 
Specifically, our \framework agents process emotions by appraising norm outcomes. 
For the emotion model, we adopt the OCC model of emotions \cite{Ortony1988OCC} in which we consider both emotional valence and intensity and assume violation of norms yields negative emotions.

\paragraph{Organization.}
The rest of the paper is structured as follows. 
Section~\ref{sec:relatedworks} discusses the relevant related works. 
Section~\ref{sec:framework} describes \framework, including the symbolic representation and the decision-making in \framework. 
Section~\ref{sec:evaluation} details the simulation experiments we conduct to evaluate \framework and describes the experimental results. 
Section~\ref{sec:discussion} presents the conclusions and the future directions. 

\section{Related Works}
\label{sec:relatedworks}

Ortony {\etal} \cite{Ortony1988OCC} model emotions based on events, action, and objects. 
Marsella and Gratch \cite{Marsella2009ema} proposed a computational model of emotion to model appraisal in perceptual, cognitive, and behavioral processes. 
Moerland {\etal} \cite{Moerland2018emotion} surveyed emotions in relation to reinforcement learning.
Keltner and Haidt \cite{Keltner1999social} differentiate the functional approaches and research of emotions by four-level analysis: individual, dyadic, groups, and cultural. Briefly, emotions provide some information for agents or people to coordinate social interactions.
We take inspiration from these works.

Savarimuthu and Cranefield \cite{Savarimuthu+Cranefield-11} proposed a life-cycles model for norms and discussed varied mechanisms of norm study. 
Broersen {\etal} \cite{Broersen2001boid} introduced the so-called Beliefs-Obligations-Intentions-Desires (BOID) architecture on top of the Beliefs-Intentions-Desires (BDI) architecture \cite{Rao1991modeling}, which further include obligation and conflict resolution.
Lima {\etal} \cite{Lima2018gavel} developed Gavel, an adaptive sanctioning enforcement framework, to choose appropriate sanctions based on different contexts.
However, these works do not consider emotions in the decision-making process.

Argente {\etal} \cite{Argente2020normative} propose an abstract  normative emotional agent architecture, which combines emotion model, normative model, and Belief-Desire-Intention (BDI) architecture. 
Argente {\etal} defined four types of relationships between emotions and norms: (1) emotion in the process of normative reasoning, (2) emotion generation with norm satisfaction or violation, (3) emotions as a way to enforce norms, (4) anticipation of emotions promotes internalization and compliance of social norms. 
Yet, Argente {\etal} do not validate the interplay between emotions and norms with their proposed architecture.

Bourgais {\etal} \cite{Bourgais2019ben} present an agent architecture that integrates cognition, emotions, emotion contagion, personality, norms, and social relations to simulate humans and ensure explainable behaviors.
However, emotions are predefined and not generated via appraisal in this work.

Von Scheve {\etal} \cite{von2006my} consider emotion generation with norm satisfaction or violation. 
Specifically, an observer agent perceives the transgression of a norm of another, its strong negative emotions (e.g., contempt, disdain, detestation, or disgust) constitute negative sanctioning of the violator. 
The negative sanctioning then leads to negative emotions (e.g., shame, guilt, or embarrassment) in the violator. 
Besides, compliance with the social norms can stem from the fear of emotional-driven sanctions, which would lead to negative emotions in the violator. 
Such fear enforces social norms. 
Yet, emotions are not part of the decision-making process in this work.

\section{\framework}
\label{sec:framework}

We now describe the architecture, norm formal model, and decision-making. 

\subsection{Architecture}
\label{sec:model}

\begin{figure}[!htb]
\centering
\includegraphics[width=0.8\textwidth]{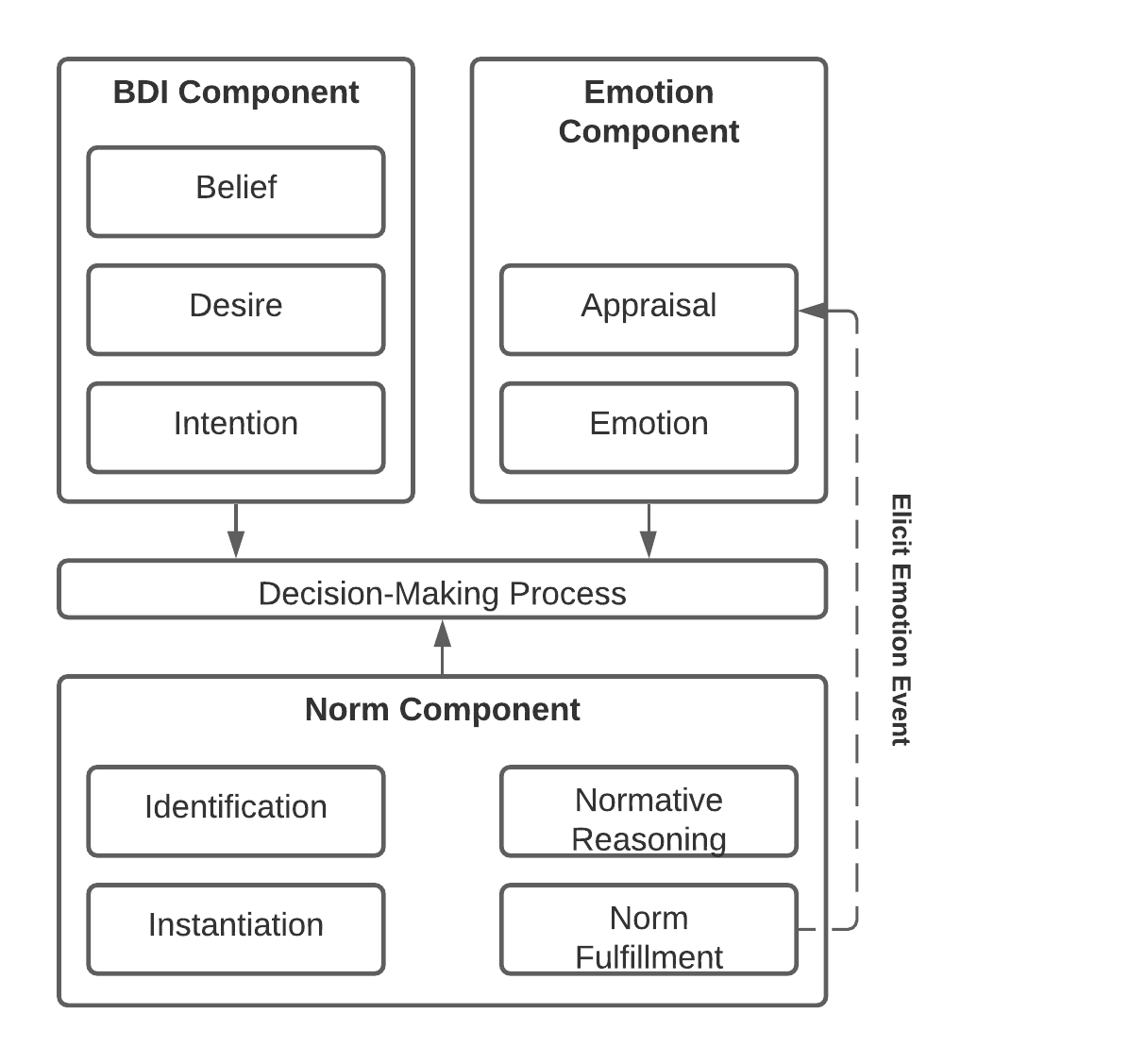}
\caption{\framework architecture, representing and reasoning over beliefs, desires, intentions, emotions, and norms.}
\label{fig:model}
\end{figure}

\framework integrates the BDI architecture \cite{Rao1991modeling} with a normative model \cite{Argente2020normative,Frantz2018modeling,Savarimuthu+Cranefield-11} and an emotional model \cite{Alfonso2017agents,Marsella2009ema}.
A \framework agent assesses the environment, including other agents' expressed emotions, its cognitive mental states, and infer possible outcomes to make a decision. Figure~\ref{fig:model} shows the three components of \framework. 

The normative component of \framework includes the following processes:
\begin{itemize}
    \item Identification: the agent recognize norms from its norm base based on its beliefs
    \item Instantiation: activate norms related to the agent
    \item Normative reasoning process: the reasoning process makes decisions based on the beliefs, current intention, self-directed emotions, other-directed emotions received from others, active norms, and how the norm satisfaction or violation influences the world and itself
    The \framework agents then update the intention based on the results of normative reasoning
    \item Norm fulfillment process: check if a norm has been fulfilled or violated based on the selected action. 
    The compliance or violation of a norm will then trigger an elicit emotion event that will be appraised at the emotion component
\end{itemize}

The BDI component includes the following parts:
\begin{itemize}
    \item Beliefs: form beliefs based on perceptions
    \item Desires: generate desires based on the beliefs
    \item Intention: the highest priority of desires to achieve based on the beliefs
    \item Action: select action based on the current intention, emotions, possible outcomes, and the evaluation of violating or complying with norms, if any

The beliefs, desires, and intentions are mental states of \framework agents.

\end{itemize}

The emotional component includes the following processes:
\begin{itemize}
    \item Appraisal: calculate the appraisal value based on the beliefs, desires, and norm satisfaction or norm violation.
    In this work, we consider only norm satisfaction or norm violation
    \item Emotion: generate emotion based on the appraisal values \cite{Marsella2009ema}
\end{itemize}

\begin{figure}[!htb]
\centering
\includegraphics[width=\textwidth]{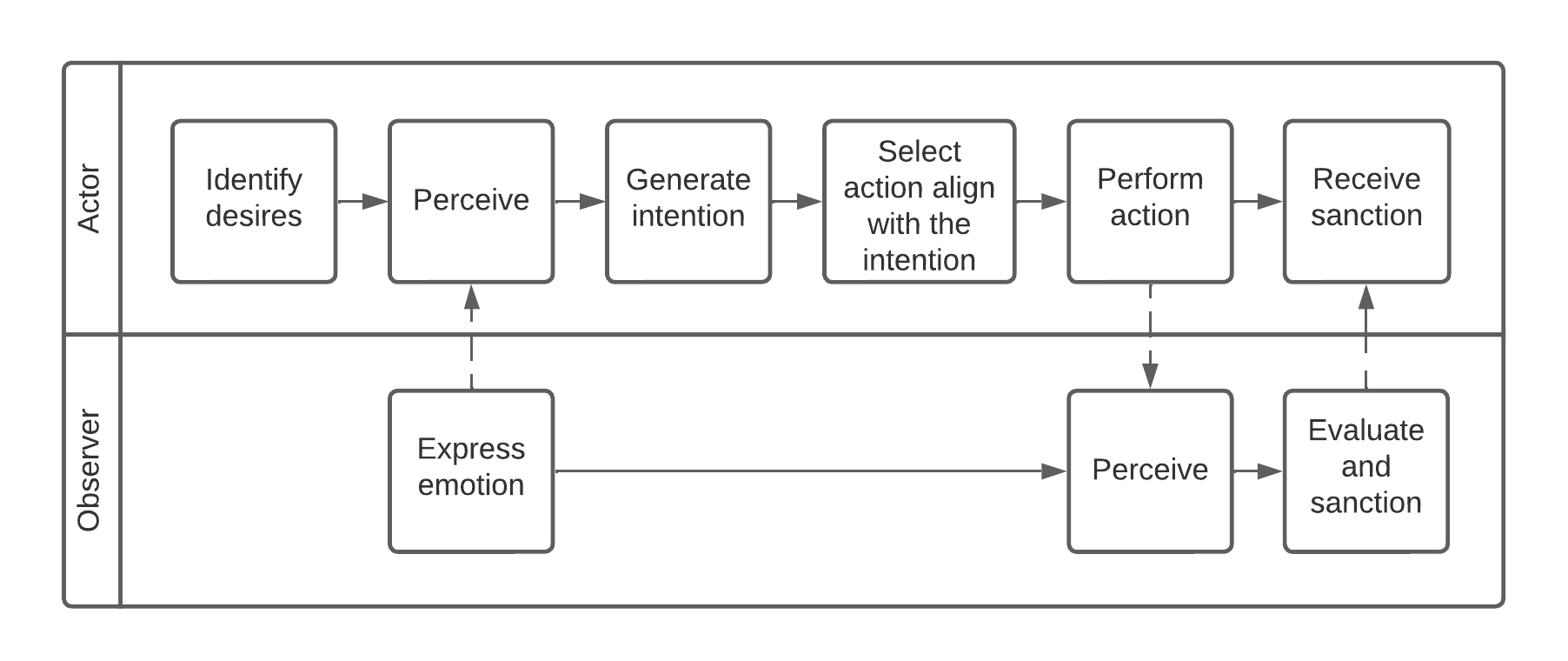}
\caption{The interaction between \framework agents.}
\label{fig:flowchart}
\end{figure}

Figure~\ref{fig:flowchart} illustrates the interactions between agents in our simulation scenario.

\subsection{Norm formal model}
\label{sec:representation}

Social norms describe the interactions between agents in a multiagent system. 
We adopt Singh's \cite{TIST-13-Governance} representation, where a social norm is formalized as $Norm(\text{subject}, \text{object}, \text{antecedent}, \text{consequent})$.
In this representation, the subject and object represent agents, and the antecedent and consequent define conditions under which the norm is activated or satisfied, respectively.
This representation describes a norm activated by the subject towards the object when the antecedent holds, and the consequent indicates if the norm was satisfied or violated. 

Following Singh \cite{TIST-13-Governance}, we consider three types of norms in \framework.
\begin{itemize}
    \item Commitment (C): the subject commits to the object to bring out the consequence if the antecedent holds.
    Consider Alice and Bob are queuing up in a grocery store. Alice and Bob commit to keeping social distance during the pandemic, represented as $C(\text{Alice}, \text{Bob}, during = \text{pandemic}, \text{social distance})$.
    \item Prohibition (P): the object prohibits the subject from the consequence if the antecedent holds.
    Caleb, the grocery store manager, prohibits Bob from jumping the queue while lining up in that store, represented as $P(\text{Bob}, \text{Caleb}, when = \text{line up}; at = \text{grocery store}, \text{jump})$.
    \item Sanction (S): same as commitment or prohibition, yet the consequence would be the sanctions. 
    Sanctions could be positive, negative, or neutral reactions to any norm satisfaction or violation
    \cite{Nardin+16:Sanctioning}.
    If Bob breaks the queue, he receives negative sanctions from Alice, represented as $S(\text{Bob}, \text{Alice}, \text{jump}, \text{negative sanctions})$. 
    Negative sanctions could be physical actions, e.g., scolding someone, or emotional expression, e.g., expressions of disdain, annoyance, or disgust.
\end{itemize}

To simulate the norm emergence and enforcement in human society,
we include emotions into the decision-making process since, by nature, humans do not always act rationally in terms of utility theory. 
Here we formalize emotions with $E_i(target, intensity, decay)$ indicating agent $a_i$ has emotion $e$ toward the target with intensity and decay value. 
An example of the prohibition case would be, Bob would not jump the queue if Alice is angry, represented as $P(\text{Bob}, \text{Alice}, \text{Bob} \succ \text{Alice} \land E_{\text{Alice}} = \text{angry}, \text{jump})$.

We model the emotional response of agents with triggered emotions from norm satisfaction, or violation \cite{Argente2020normative}. 
Here we represent the elicited emotions with $Elem_{name}(A_{expect}, A_{real}, Em_1, Em_2) | Em_1, Em_2 \in E; A_{expect}, A_{real} \in A$ where A is a set of actions. 
E is a set of emotions, and $Em_1$ and $Em_2$ are the emotions triggered by norm satisfaction and violation accordingly. 
If the $A_{expect}$ is equal to the $A_{real}$, a norm has been fulfilled, and $Em_1$ was elicited. 
$Ap(beliefs, desires, Elem)$ represents the appraisal function.

\subsection{Decision-Making}
\label{sec:decision-making}
Schwarz \cite{Schwarz2000emotion} addresses the influence of moods and emotions at decision making and discusses the interplay of emotion, cognition, and decision making. 
Specifically, the aspects include pre-decision affect, post-decision affect, anticipated affect, and memories of past affect. 
In our model, we include the pre-decision affect into the decision-making process. 
With pre-decision affect, people recall information from memories that match their current affect \cite{Schwarz2000emotion}. 
\revised{For instance, people in a sad emotion or interacting with hostile people tend to overestimate adverse outcomes and events.}

In our model, emotions serve as mental objects \revised{and an approach to sanctioning}.
\revised{We consider emotions as intrinsic rewards from agents' internal state in contrast to physical rewards from the environment.}
\revised{We adopt the OCC model of emotions \cite{Ortony1988OCC}, in which we consider emotional valence and intensity.}
We formulate emotions with simple values where positive values indicate positive emotions and larger values indicate higher intensity.
\revised{A mood is a general feeling and not a response to a specific event or stimulus compared to emotions.
Therefore, we consider emotions but not mood.}
\framework agents' appraisal function considers norm satisfaction and violation only. 
The agents are aware of other agents' expressed emotions in the same place.
In this work, we assume that agents express true and honest emotions and can correctly perceive the expressed emotions. In other words, felt emotions are equal to expressed emotions. 
Another assumption is that emotions are consistent with the notions of rational behavior.

Algorithm~\ref{decisionloop} displays the decision loop of our model.
At the beginning of the simulation, all agents are initialized with certain desires, and during the run, an intention would be generated by prioritizing desires with the agent's beliefs. 
When choosing the next move with line 5 in Algorithm~\ref{decisionloop}, the agent chooses the one with maximum utility from all available actions.
Algorithm~\ref{action_selection} details the action selection.
The decision takes the agent's beliefs, current intention, and possible consequences into accounts. 
While norms are activated with the beliefs, the agent would further consider emotions and cost and possible consequences with norms at line 9 in Algorithm~\ref{action_selection}. 
For instance, if people violate some social norms, they may be isolated from society. 
Regarding the influence of emotions, people may overestimate the negative outcomes when they are in the negative emotion and tend to comply with the norms. 

\begin{algorithm}[!ht]
\SetAlgoLined
    Initialize one agent with its desires D\;
    \For{t=1,T}{
        Observe the environment (including the expressed emotions from others $E_{around}$) and form beliefs $b_t$\;
        Generate intention I based on $b_t$ and D\;
       
        $a_t$ = ActionSelection($b_t$, I, D)\;
        Execute action $a_t$\;
        
        Elicit self-directed emotions $E_{self}$ from agent itself based on if action $a_t$ fulfills a norm\;
        Self-sanction with $E_{self}$\;
        
        Observe the environment (including the performed actions $a_{t\_other}$ of other agents) and form beliefs $b_{t+1}$\;
        
        Elicit other-directed emotions $E_{other}$ for observer agents based on if action $a_{t\_other}$ fulfills a norm\;
        
        Sanction others with $E_{other}$\;
        
    }
 \caption{Decision loop of a \framework agent}
 \label{decisionloop}
\end{algorithm}

\begin{algorithm}[!ht]
\SetAlgoLined
\KwInput{beliefs $b_t$, intention I, desires D}
\KwOutput{Action $a_t$}
\SetKwProg{Fn}{Function}{:}{\KwRet}
\Fn{\FMain}{
    $E_{around} \subset b_t$\;
    \For{each a in ACTIONS($b_t$)}
    {
        Activate norms N with beliefs $b_t$ and a\;
        \eIf{N $= \varnothing$}{
            $a_t$ = MAX$_a$(RESULT($b_t$, intention, a))
        }{
            $a_t$ = MAX$_a$(RESULT($b_t$, intention, a, N) $\times$ amplifier($E_{around}$) )
        }
    }
    
    \KwRet $a_t$
}

\caption{Action selection}
\label{action_selection}
\end{algorithm}

\section{Evaluation}
\label{sec:evaluation}

We evaluate \framework via a line-up environment where agents form queues to receive service. We detail the environment in Section~\ref{sec:environment}.

\subsection{Line-up Environment}
\label{sec:environment}

\begin{figure}[!htb]
\centering
\includegraphics[width=\textwidth]{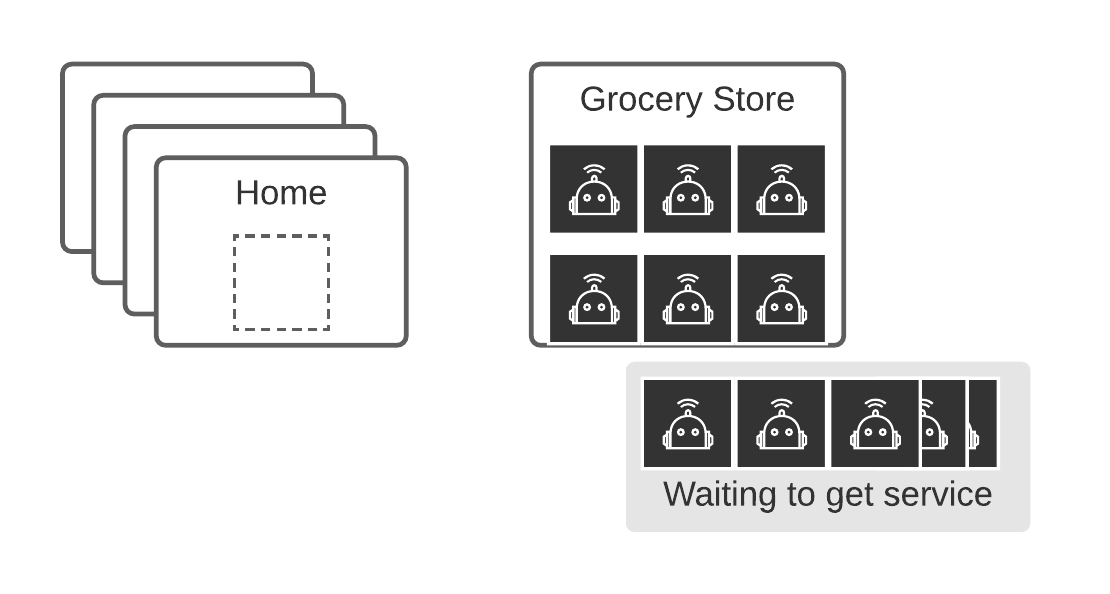}
\caption{Simulation details.
Agents move between their homes and the grocery store.
The store has a capacity limit of eight customers at one time.
As a result, other agents must line up outside the store to get service.}
\label{fig:location}
\end{figure}

Figure~\ref{fig:location} shows the line-up environment. 
We build this line-up environment using Mesa \cite{Masad2015mesa}, a Python-based framework for building, analyzing, and visualizing agent-based models. 

The line-up environment includes two shared locations---home and grocery stores.
The agents move between home and grocery stores to get food.
We consider one social norm in the line-up environment: agents are expected to line up to enter the grocery store. 
To simulate real human reactions to norm violations, we refer to a social psychology experiment \cite{Milgram1986response}. In the line-up environment, we model defensive reactions of people in the queue as negative emotions toward those who jump the queue by barging in ahead of someone already in the queue. Conversely, people show positive emotions toward those who stay in the queue. 

We initialize the agents with the following parameter values:

\begin{itemize}
    \item Health (Integer value from 0--100): When the health value reaches zero, the agent is marked as \fsl{deceased} and unable to act. The health value decreases by 1 unit at each step. 
    \item Deceased (Boolean: True or False): set as True when an agent runs out of health. 
    \item Emotion (Integer value): simplified with numerical values where positive values indicate positive emotions and negative indicate negative emotions. The emotions come along with a duration. Default at 0.
    \item Number of food packets owned (Integer value from 0--15): once obtained food from the stores, agents would be able to restore its health value via consuming food anywhere.
    \item Food expiration day (Integer value from 0--15): once the agent gets food packets, we update the expiration day with 15. The expiration day decreases by 1 unit at each step. Food expires once the expiration day reaches 0.  Default at 0. 
    \item Beliefs: the perceived and processed information from the world, including other agents' expressed emotions.
    \item Desires: desired states, including \fsl{have food} and \fsl{wandering}.
    \item Intention: the highest priority of desires to achieve at a specific time. When the agent's health is lower than the threshold, 80\% of the health, the agent sets its intention as \fsl{get food}; otherwise, the agent sets its intention as \fsl{wandering}.
\end{itemize}

When an agent runs low on stock, it has a higher probability of moving to a grocery store. The grocery store can provide food packets to eight agents in one time step. 
While waiting in line to get food, the agent could either stay in the line or jump ahead in the line to get food in less time. 
Jumping the line may increase other agents' delay in getting food packets. 
Those who witness the violation would then cast negative emotions, further interpreted as anger or disdain, triggered by that behavior. 
To simplify the simulation, we presume the anticipated affects \cite{Schwarz2000emotion} with: (1) receiving negative emotions triggers negative self-directed emotions such as shame and guilt;  (2) complying with norms leads to positive or neutral emotions; (3) violating norms leads to negative or neutral emotions. 
The intensity of emotions triggered each time is fixed, but the values of emotions can add up.
Each triggered emotion lasts 2 steps. 
At each step, the duration and intensity of emotion decrease by 1 as decay.
A simple assumption here is that people in a bad mood would trigger stronger emotions in response to a non-ideal state.
Note that at the beginning of the simulation, we initialize the agent society with health in normal distribution to avoid all agents having the same intention at the same time.

\subsection{Agent Types}

To answer our research question and evaluate \framework, we define three agent societies as baselines. 
We describe the agents societies below: 

\begin{description}
\item[Obedient society.] Agents in an obedient society always follow norms. 
\item[Anarchy society.] Agents in an anarchy society jump lines when they cannot get food.
\item[Sanctioning society.] Agents in the sanctioning society jump lines considering the previous experience of satisfying or violating a norm.
Agents sanction positively or negatively based on norm satisfaction or violations directly and comply with enforced norms.

\item [\framework society.] Agents in the \framework society jump lines considering the previous experiences of satisfying or violating a norm, current emotional state of the other agents, current self emotional state, and estimated outcome of satisfying or violating a norm. 
\framework agents who observe norm satisfaction or violations would appraise the norm outcomes and trigger emotions to sanction the actor agent.
\end{description}

Table~\ref{tab:society_characteriscs} summarizes the characteristics of the agents in the four societies. 

\begin{table}[!htb]
\centering
\caption{Characteristics of the various agent societies.}
\label{tab:society_characteriscs}
    \begin{tabular}{l ccc}
        \toprule
        Agent Type & Violation allowed & Sanctioning & Emotions involved \\\cmidrule(r){1-1} \cmidrule(lr){2-2} \cmidrule(lr){3-3} \cmidrule(l){4-4}
        Obedient society    & \ding{56} & \ding{56} & \ding{56} \\
        Anarchy society     & \ding{52} & \ding{56} & \ding{56} \\
        Sanctioning society & \ding{52} & \ding{52} & \ding{56} \\
        \framework society  & \ding{52} & \ding{52} & \ding{52} \\
        \bottomrule
    \end{tabular}


\end{table}

\subsection{Hypotheses and Metrics}

To address our research question RQ\fsub{emotion} on emotions and norm emergence, we propose three hypotheses:

\begin{description}
    
    \item[H\fsub{1} (Norm satisfaction):] Norm satisfaction in \framework agent society is higher compared to the baseline agent societies.
    \item[H\fsub{2} (Social welfare):] \framework agent society yields better social welfare compared to the baseline agent societies.
    \item[H\fsub{3} (Social experience):] \framework agent society yields a better social experience compared to the baseline agent societies.
\end{description} 

\revised{To evaluate H\fsub{1} on norm satisfaction, we compute one metric, M\fsub{1} (Cohesion): Percentage of norm satisfaction.}

To evaluate H\fsub{2} on social welfare, we compute two metrics:
\revised{ (1) M\fsub{2} (Deceased): Cumulative number of agents deceased; (2) M\fsub{3} (Health): Average health of the agents.}

\revised{To evaluate H\fsub{3} on social experience, we compute one metric, M\fsub{4} (Waiting time): Average waiting time of agents in the queues.}

To test the statistical significance of H\fsub{1}, H\fsub{2}, and H\fsub{3}, we conduct the independent t-test and measure effect size with Glass's $\Delta$ for unrelated societies \cite{Grissom2012effectsizes,Glass1976primary}. We adopt Cohen's \cite[pp.~24--27]{Cohen-88-Statistics} descriptors to interpret effect size where above 0.2, 0.5, 0.8 indicate small, medium, and large.

\subsection{Experimental Setup}
\label{sec:experiment-setup}

We run each simulation with 400 agents and queue size 80 for 3,000 steps.
We choose a relatively small number of agents to reduce the simulation time while our results are stable for a more significant number of agents.  
The simulation stabilizes at about 1,500 steps, but we keep extended simulation steps to have more promising results. 
\revised{Table~\ref{tbl:payoffs} lists the payoffs applied in our simulation.}

We present the results with a moving average of 100 steps.
We choose this size of running window to show the temporal behavior change in a small sequence of time. 
With a larger size, the running window may alleviate the behavior change.
To minimize deviation from coincidence, we run each simulation with 10 iterations and compute the mean values.

\begin{table}[!htb]
\centering
\caption{Payoff table.}
\label{tbl:payoffs}
\begin{tabular}{l l r}\toprule
Component & Type & {Reward} \\\midrule
Deceased & Extrinsic & --500 \\
Norm compliance \& positive emotion & Intrinsic & 1 \\
Norm violation \& negative emotion & Intrinsic & --1 \\
\bottomrule
\end{tabular}
\end{table}

\subsection{Experimental Results}
\label{sec:results}

In this section, we describe the simulation results comparing the three baselines and \framework agents. 
Table~\ref{tab:results} summarizes these results. 
Table~\ref{tab:stats} lists the value of Glass's $\Delta$ and $p$-values from the independent t-test.

According to Table~\ref{tab:results} and Table~\ref{tab:stats}, we see that \framework generate better cohesion and fewer deceased agents than baselines (p $< 0.01$; Glass's $\Delta > 0.8$). 
The null hypothesis corresponding to H\fsub{1} is rejected. 
Note that we do not consider the cohesion metric for the obedient agent society here since agents in the obedient society are always compliant.
However, \framework also yields the worst social experience where the low waiting time is a desirable state (p $< 0.01$; Glass's $\Delta > 0.8$).

\begin{table}[!htb]
\centering
\caption{Comparing \framework agent society with baseline agent societies on various metrics.}
\label{tab:results}    
    \begin{tabular}{lrrrrr}
        \toprule 
        Agent Society & Cohesion & Deceased & Health & Waiting Time \\ \midrule
        Obedient & $\shyphen$ & 55.30 & 79.27 & 8.95 \\
        Anarchy & 0.22 & 81.60 & 79.50 & 5.45 \\
        Sanctioning & 0.88 & 169.30 & 86.26 & 2.55 \\
        \framework & \textbf{0.99} & \textbf{54.00} & 79.00 & 8.95 \\
        \bottomrule
    \end{tabular}

\end{table}

\begin{table}[!htb]
\centering
\caption{Statistical analysis.}
\label{tab:stats}    
    \begin{tabular}{l rrrr rrrr}
        \toprule 
        \multirow{2}{*}{Agent Society}
        &
        \multicolumn{4}{c}{Glass's $\Delta$} &
        \multicolumn{4}{c}{$p$-value} \\\cmidrule(lr){2-5}\cmidrule(l){6-9}
         & Cohesion & Deceased & Waiting time & Health &
        Cohesion & Deceased & Waiting time & Health \\ 
        \midrule
        Obedient & 0.19 & 0.65 & 0.01 & 0.18 & 0.32 & $<0.01$ & 0.98 & 0.52 \\
        Anarchy & 102.43 & 3.10 & 40.82 & 0.21 & $<0.01$ & $<0.01$ & $<0.01$ & 0.46 \\
        Sanctioning & 13.67 & 15.53 & 76.68 & 3.34 & $<0.01$ & $<0.01$ & $<0.01$ & 8.45 \\
        \framework & -- & -- & -- & -- & -- & -- & -- & -- \\
        \bottomrule
    \end{tabular}

\end{table}

\subsubsection{H\fsub{1} Norm Satisfaction}

\begin{figure}[!htb]
\centering
\includegraphics[width=0.6\textwidth]{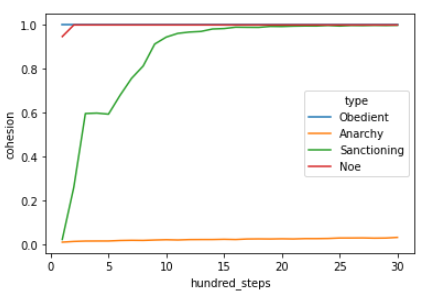}
\caption{Simulation result: average cohesion. Comparing average cohesion (M\fsub{1}) yielded by \framework and baseline agent societies.}
\label{fig:cohesion}
\end{figure}

Figure~\ref{fig:cohesion} displays the cohesion, the percentage of norm satisfaction, in the baseline agent societies and the \framework agent society. 
We find that the percentage of norm satisfaction in the \framework agent society, average at 99\% and p-value $<0.01$, is constantly higher than the sanctioning agent society, average at 88\% and p-value $<0.01$ and Glass's $\Delta > 0.8$. 
The sanctioning agent society learns to comply with the norm as time goes by. The \framework agent society does sanction as well. 
Yet, considering emotions and the possible outcome makes \framework agent society enforce the norm faster than the sanctioning agent society. 
Specifically, \framework agent society enforces the norm at about 100 steps while sanctioning agent society at \np{1500} steps.

\subsubsection{H\fsub{2} Social Welfare}

\begin{figure}[!htb]
\centering
\includegraphics[width=0.6\textwidth]{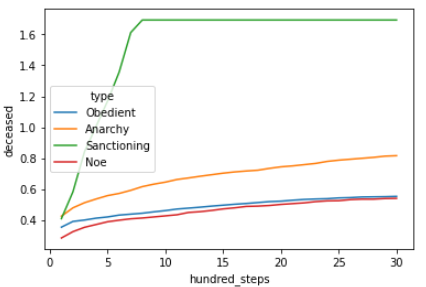}
\caption{Simulation result: average number of deceased.Comparing average number of deceased (M\fsub{2}) in \framework and baseline agent societies.}
\label{fig:deceased}
\end{figure}

Figure~\ref{fig:deceased} compares the average number of deceased in the obedient, anarchy, sanctioning, and \framework agent societies. 
Refer to Figure~\ref{fig:cohesion}, sanctioning agent society learns the norm via positive and negative sanctioning from norm satisfaction and violation. 
However, the agents in that society do not consider the possible severe consequences and cause compliant agents to die in the queue. 
When the number of deceased reaches the threshold, the simulation stabilizes. Therefore, no more agent from the sanctioning agent society dies after the threshold. On the contrary, \framework agent society sanctions and considers possible outcomes of norm satisfaction and violation, therefore learning the norm and avoiding unacceptable consequences.

\begin{figure}[!htb]
\centering
\includegraphics[width=0.6\textwidth]{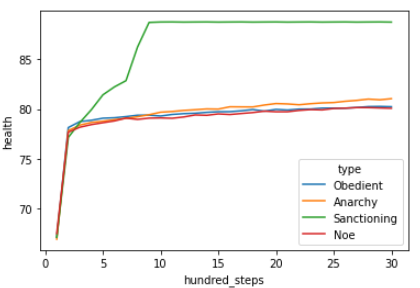}
\caption{Simulation result: average health value. Comparing average health value (M\fsub{3}) in \framework and baseline agent societies.}
\label{fig:health}
\end{figure}

Figure~\ref{fig:health} compares the average health of the agents in the obedient, anarchy, sanctioning, and \framework agent societies. 
The sanctioning agent society yields higher health State, with a mean at 86.26, but at the expense of more deaths.
The rest of the agents then be able to remain in high health.

\subsubsection{H\fsub{3} Social Experience}

\begin{figure}[!htb]
\centering
\includegraphics[width=0.6\textwidth]{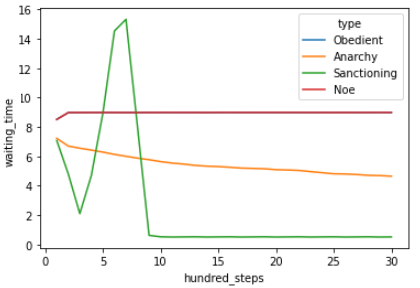}
\caption{Simulation result: average waiting time of agents in queues. Comparing average waiting time (M\fsub{4}) in \framework and baseline agent societies.}
\label{fig:duration}
\end{figure}

Figure~\ref{fig:duration} compares the average waiting time the agents spend in a queue at the grocery store in the obedient, anarchy, sanctioning, and \framework agent societies. 
The \framework agent society learns the norm fast and remains the same waiting time in the queue. 
However, some agents in the sanctioning agent society take advantage of those who learn norms faster than themselves. 
Therefore, many agents die during the learning process, and the simulation stabilizes. 
In Figure~\ref{fig:duration}, the obedient agent society shares the same trend with \framework agent society since emotions enforce the line-up norm.

Combining the results for H\fsub{1} and H\fsub{2} and H\fsub{3}, we note that while sanctioning enforces norms, a combination of sanctioning and emotions enforce norms better. 
Specifically, having emotions as amplifiers of outcomes yield higher norm satisfaction compared to our baselines. 
The results also indicate that, first, sanctioning agents that consider only norm violation or norm satisfaction may bring out worse social welfare compared to \framework that considers both norms and their consequences. 
Second, although \framework agents remain relatively high waiting time in the queues, the number of deceased is lower than the baselines. 
Note that the sudden drop of deceased number or increase of health value for sanctioning agents resulted from the stabilization of that society. 
Third, \framework agents stay in positive emotions during the simulation while sanctioning agents start from negative emotions and eventually achieve the expected behaviors.

\section{Discussion and Conclusion}
\label{sec:discussion}

We present an agent architecture inspired by the norm life-cycle \cite{Argente2020normative}, BDI architecture \cite{Rao1991modeling}, and emotion life-cycle \cite{Alfonso2017agents,Marsella2009ema} to investigate how emotions influence norm emergence and social welfare.
We evaluate the proposed architecture via simulation experiments in an environment where agents queue up to receive service. 
Our simulations consider two characteristics of an agent society: sanctioning and emotions that participate in action selection and arise from evaluating selected action. 
The experiments show that incorporating emotions enables agents to cooperate better than those who do not.

In our agent architecture, we make an assumption that agents can recognize others' emotions. However, we acknowledge that emotion recognition is a challenging task \cite{Barrett2019emotional}. Whereas recent works in AI have focused on emotion recognition through facial expressions and emotion recognition using wearables, it is worth noting that there is no agreement in modeling emotions in the psychology community \cite{Barrett2019emotional,Marin2018affective,Marsella2010computational}.

Murukannaiah {\etal} \cite{AAMAS-20:Blue-Sky} address many shortcomings of current approaches for AI ethics, including taking the value preferences of an agent's stakeholder and other agents' users, learning value preferences by observing the responses of other agents' users, and value-based negotiation.
Incorporating these aspects in \framework is an interesting future direction.

As a future extension of current work, we plan to differentiate emotions in \framework instead of \revised{modeling emotions with emotion valences} to provide more information for value preferences. 
We also consider including a mix of personalities in future research to have different appraisal results.
In this work, \framework agents are assumed to express true and honest emotions. 
However, emotions can also serve as a tool to influence, persuade, or deceive others in an adversarial context.
It would be crucial to identify and model these contradictions while humans are in the loop.

\section*{Acknowledgments}
STT and MPS thank the NSF for partial support under grant IIS-1908374.

\bibliographystyle{splncs04}

\bibliography{Munindar,Nirav,SzTing}

\end{document}